\documentclass[letterpaper]{article} 
\usepackage{aaai2026}  
\usepackage{times}  
\usepackage{helvet}  
\usepackage{courier}  
\usepackage[hyphens]{url}  
\usepackage{graphicx} 
\usepackage{xcolor}
\usepackage{color}
\usepackage{natbib}  
\usepackage{caption} 
\usepackage{amsmath}
\usepackage{amssymb}
\usepackage{pifont}
\usepackage{multirow}
\usepackage{booktabs}
\usepackage{algorithm}
\usepackage{algorithmic}

\usepackage{newfloat}
\usepackage{listings}
\DeclareCaptionStyle{ruled}{labelfont=normalfont,labelsep=colon,strut=off} 
\floatstyle{ruled}
\newfloat{listing}{tb}{lst}{}
\floatname{listing}{Listing}
\makeatletter
\def\thanks#1{\protected@xdef\@thanks{\@thanks
        \protect\footnotetext{#1}}}
\makeatother

\title{MODA: The First Challenging Benchmark for Multispectral \\  Object Detection in Aerial Images}
\author{
    Shuaihao Han, Tingfa Xu\textsuperscript{*}, Peifu Liu, Jianan Li\textsuperscript{*}\thanks{\textsuperscript{*}Correspondence to: Tingfa Xu and Jianan Li.}
}
\affiliations{

    Beijing Institute of Technology
    
}

\usepackage{bibentry}

\begin{document}
 
\maketitle

\begin{abstract}
Aerial object detection faces significant challenges in real-world scenarios, such as small objects and extensive background interference, which limit the performance of RGB-based detectors with insufficient discriminative information. Multispectral images (MSIs) capture additional spectral cues across multiple bands, offering a promising alternative. However, the lack of training data has been the primary bottleneck to exploiting the potential of MSIs. To address this gap, we introduce the first large-scale dataset for Multispectral Object Detection in Aerial images (MODA), which comprises 14,041 MSIs and 330,191 annotations across diverse, challenging scenarios, providing a comprehensive data foundation for this field. Furthermore, to overcome challenges inherent to aerial object detection using MSIs, we propose OSSDet, a framework that integrates spectral and spatial information with object-aware cues. OSSDet employs a cascaded spectral-spatial modulation structure to optimize target perception, aggregates spectrally related features by exploiting spectral similarities to reinforce intra-object correlations, and suppresses irrelevant background via object-aware masking. Moreover, cross-spectral attention further refines object-related representations under explicit object-aware guidance. Extensive experiments demonstrate that OSSDet outperforms existing methods with comparable parameters and efficiency. 
\end{abstract}

\begin{links}
    \link{Datasets}{https://github.com/shuaihao-han/MODA}
\end{links}

\section{Introduction}
\begin{figure}[t]
    \centering
    \includegraphics[width=\linewidth]{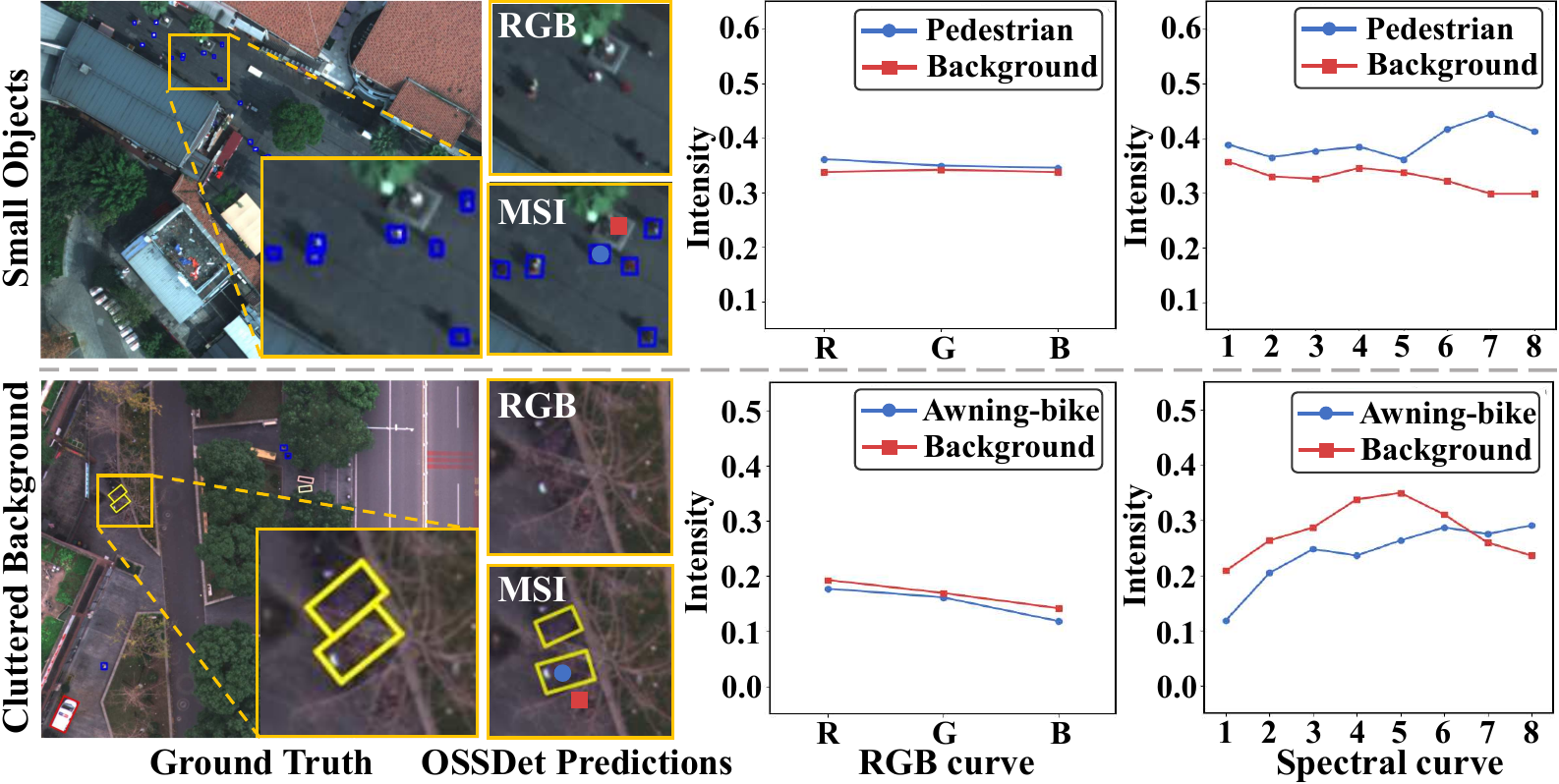}
    \caption{In challenging scenarios, limited spatial information in RGB data hampers effective detection. In contrast, multispectral images offer additional spectral cues that significantly enhance target discrimination for robust detection.}
    \label{fig:fig1}
\end{figure}

Aerial object detection has become a prominent research focus in recent years~\cite{VTUAV, xie2021oriented}, yet remains challenging in practice due to small targets and strong background noise~\cite{leng2024recent}, which obscure target features and degrade detection performance. This issue is particularly pronounced for RGB-based detectors that rely heavily on spatial cues (Fig.~\ref{fig:fig1}).

Multispectral images (MSIs) capture additional spectral cues that characterize the target’s intrinsic reflectance properties~\cite{fang2023toward}, offering valuable insights in cases where spatial features are limited by aforementioned challenges~\cite{fang2023hyperspectral}. For example, distinct spectral curves reliably differentiate targets from background, even in cases involving small objects or cluttered scenes (Fig.~\ref{fig:fig1}). Moreover, critical spectral features generally remain stable despite variations in object appearance~\cite{qin2024dmssn}, offering robust cues for detection. Consequently, MSIs represent a promising approach for object detection in challenging aerial images. However, the limited availability of large-scale datasets hampers further advancement.

To bridge this gap, we present MODA, the first large-scale challenging dataset for multispectral aerial object detection, containing 14,041 MSIs with 330,191 oriented annotations across 8 categories. Each MSI features a large image size of 1200$\times$900 and covers 8 spectral bands (395$\sim$950nm). The data is captured in diverse urban scenes over various time periods. Additionally, MODA includes 8 challenging attributes commonly encountered in real-world scenarios (e.g., small objects, low illumination, truncation, occlusion). This comprehensive design renders MODA well-suited for practical applications and establishes it as a valuable resource for advancing research in multispectral aerial object detection.

In terms of model design, leveraging spectral-spatial information of MSI for aerial object detection encounters three key challenges: (i) prior multispectral object detection (MOD) methods decouple spectral and spatial information via dimensionality reduction such as PCA~\cite{pearson1901liii} and band selection~\cite{zhang2024tensorial} along with two-stream network for independent processing, incur high computational costs and spectral information loss; (ii) existing aerial object detection methods primary exploiting spatial features~\cite{xiao2024lightweight}, and spectral cues is underutilized; (iii) extensive background interference in complex scenarios dilute distinctive target features~\cite{visdrone}, impairs model attention and object spectral-spatial feature learning, degrade detector performance.

To address these issues, we propose OSSDet, an object-aware spectral-spatial learning framework for multispectral aerial object detection. OSSDet integrates spectral and spatial cues in a unified object-aware design, eliminating explicit decoupling. It employs a cascaded spectral-spatial awareness interactive modulation structure to optimize target perception and a Spectral-Guided Adaptive Cross-Layer Fusion module that adaptively aggregates spectral-related features to reinforce intra-object correlations, resisting noise interference while preserving spatial texture details to minimize information loss during cross-layer fusion.

Moreover, to mitigate significant background noise that dilutes target features, especially for small objects, we introduce an object-aware masking branch. This branch generates an object activation mask from low-level features using an object activation loss as guidance. As an object-aware prior, the mask refines these features by preserving object-specific information while suppressing irrelevant noise. Subsequently, features with explicit object-aware cues are integrated with cross-spectral attention to bolster object-related representations in both spatial and spectral domains. 

Extensive experiments on MODA and other MOD datasets demonstrate that OSSDet achieves state-of-the-art performance while preserving computational efficiency. The primary contributions of this work are summarized below:
\begin{itemize}
    \item We introduce the first large-scale dataset for multispectral aerial object detection, featuring diverse challenges to support research advancements in this domain.
    \item We propose OSSDet, a novel framework that integrates spectral and spatial information with object-aware cues for robust detection in aerial scenarios using MSI.
    \item We develop cascaded spectral-spatial awareness modulation to enhance target perception, spectral aggregation to reinforce intra-object correlations, and object-aware masking branch to reduce noise interference, optimizing MSI utilization and improving detection accuracy.
\end{itemize}

\section{Related Work}
\textbf{Multispectral Object Detection Datasets.} 
Growing interest in multispectral object detection has led to the introduction of several datasets. HOD-1~\cite{yan2021object} is the first bounding box-annotated MSI dataset, followed by HOD3K~\cite{he2023object}, which captures MSIs in natural scenes. However, HOD-1 includes manually placed objects with simple fore-backgrounds, and HOD3K relies on fixed camera positions, resulting in static backgrounds that lack real-world dynamics. To address these limitations, we propose MODA, a large-scale dataset with diverse real-world challenges for advancing research in multispectral aerial object detection. To our best knowledge, MODA is the largest multispectral object detection dataset available.

\noindent\textbf{Aerial Object Detection.} 
Many existing methods address the boundary discontinuity issue in oriented object angle prediction~\cite{yang2021rethinking, xu2024rethinking}, improve sample quality and allocation~\cite{oriented-reppoints}, or enhance object representations~\cite{han2021align, Li_2024_IJCV, yuan2025strip}. However, these approaches are confined to RGB images, overlooking the rich spectral cues in MSIs. To bridge this gap, we propose a novel framework that exploits spectral information to advance multispectral aerial object detection.

\noindent\textbf{Multispectral Object Detection.} 
Despite advances in general object detection \cite{carion2020end, tian2025yolov12}, MOD remains underexplored due to limited data. \cite{yan2021object} pioneered this area with the first dataset and a 3D CNN for spatial-spectral feature extraction, achieving promising results. More recently, \cite{he2023object} proposed S2ADet, a two-stream network separating spatial and spectral branches for independent feature extraction and fusion, improving accuracy while incurring high computational cost due to the decoupling and two-stream network complexity. In contrast, our method adopts a single-stream design, integrating spectral and spatial information directly, achieves superior performance while maintaining computational efficiency, and avoids the limitations of prior methods.
\begin{table*}[t]
\centering
    {\small
    \setlength{\tabcolsep}{2.0pt}
    \begin{tabular}{l|cccccccccccc}
    \toprule
    \textbf{Dataset} & \textbf{Scene} & \textbf{Images} & \textbf{Image size} & \textbf{Bands} & \textbf{Categories} & \textbf{Annotations} & \textbf{Avg.labels/image} & \textbf{Spectral bands} & \textbf{Type} & \textbf{Year} \\
    \midrule
    HOD-1~\cite{yan2021object} & Manual & 454 & 467$\times$336 & 96 & 8 & 1657 & 3.57 & 400$\sim$1000nm & HBB & 2021 \\
    HOD3K~\cite{he2023object}  & Natural &3242 & 512$\times$256 & 16 & 3 & 15149 & 4.37 & 470$\sim$620nm & HBB & 2023 \\
    \midrule
   \textbf{MODA (Ours)} & Aerial & 14041 & 1200$\times$900 & 8 & 8 & 330191 & 23.52 & 390$\sim$950nm & OBB & 2025 \\
    \bottomrule
    \end{tabular}
    }
\caption{Comparison of multispectral object detection datasets. HBB: horizontal bounding box; OBB: oriented bounding box.}
\label{tab1}
\end{table*}

\section{MODA Dataset}
\subsection{Overview}
\noindent \textbf{Data Collection.} 
Multispectral sensors trade spatial resolution for spectral bands. MODA targets aerial scenes with widespread small objects requiring high spatial resolution; thus, we select a professional drone-mounted multispectral camera (1280$\times$960 image size; 8 spectral bands range of 395$\sim$950 nm; 4.5cm/pixel at 100m height) to record MSIs across diverse scenes, times, and weather, yielding 14,041 MSIs (9,156 training; 4,885 testing) across 50 urban areas.

\noindent \textbf{High Quality Annotation.} 
For high annotation quality, we adopted a three-stage protocol: (i) drafting detailed guidelines (object definitions, annotation tool usage) and training annotators via trial tasks; (ii) trained annotators labeled the raw data accordingly; (iii) two verification passes by the author team to correct errors and ensure accuracy.

\begin{figure}[t]
    \centering
    \includegraphics[width=\linewidth]{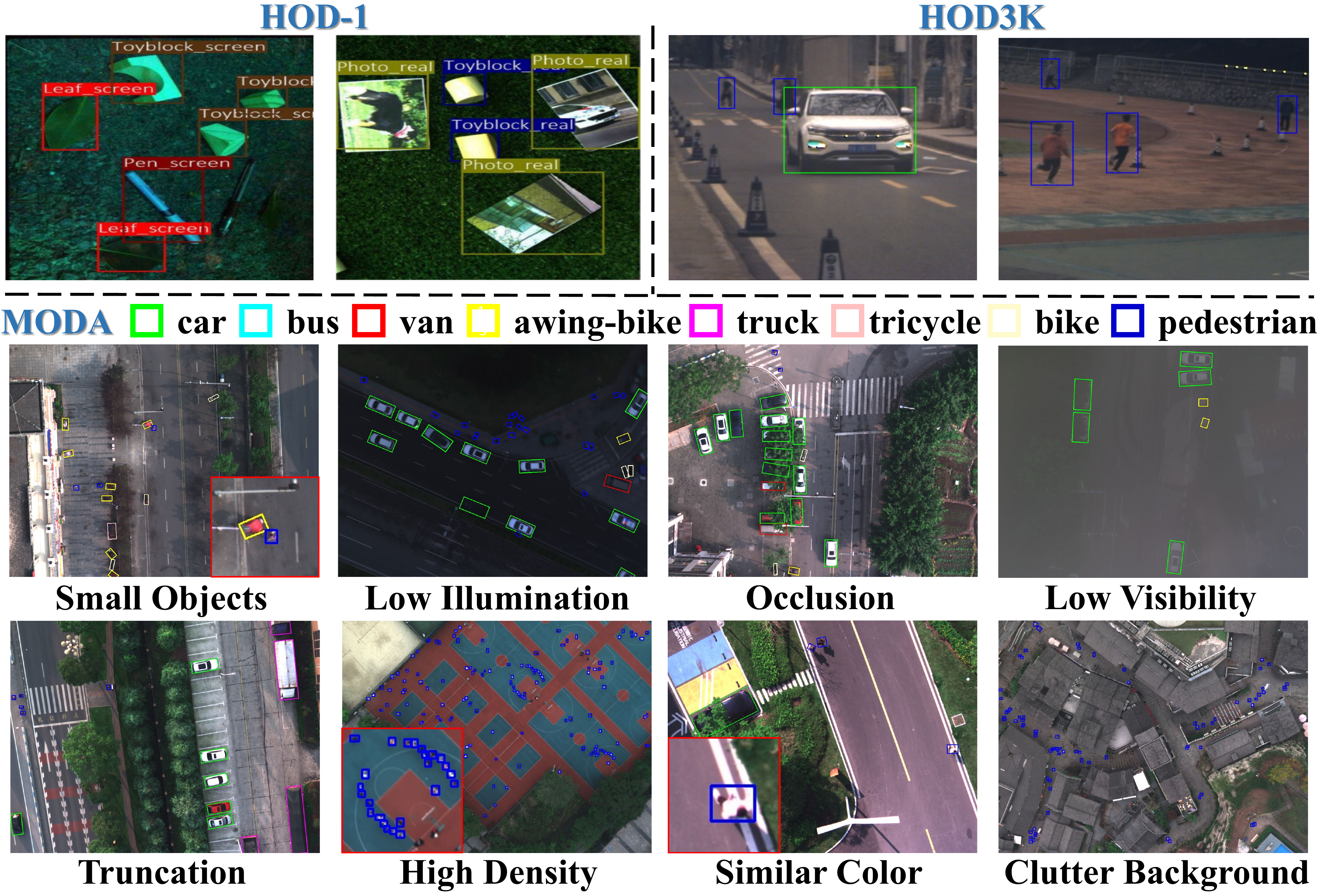}
    \caption{Comparison with other multispectral object detection datasets and examples of challenge attributes in MODA.}
    \label{fig:challenge_attributes}
\end{figure}

\noindent \textbf{Dataset Comparison.} 
Existing MOD datasets suffer from limited scale and artificial ground scenarios (Table~\ref{tab1}). HOD-1 targets manually placed targets, while HOD3K features fixed scenes with static backgrounds, limiting their generalization (Fig.~\ref{fig:challenge_attributes}). In contrast, MODA offers 14,041 MSIs with 330,191 targets across 8 categories using oriented bounding boxes. Collected from diverse aerial views, MODA greatly expands data scale and complexity, better reflecting real-world challenges to support broader practical applications.

\subsection{Dataset Analysis}
\textbf{Challenge Attributes.} 
As Fig.~\ref{fig:challenge_attributes} shows, HOD-1 targets manually placed objects captured from real scenes or screens, and HOD3K focuses on occlusion and small objects in natural scenes. MODA targets to tackle more complex aerial scenarios by integrating spectral and spatial information. We define and record eight key challenge attributes of aerial object detection, including small objects, occlusion, and low visibility (Fig.~\ref{fig:challenge_attributes}), which hinder detectors that rely solely on RGB data, underscoring the need for spectral cues.

\begin{figure}[t]
    \centering
    \includegraphics[width=\linewidth]{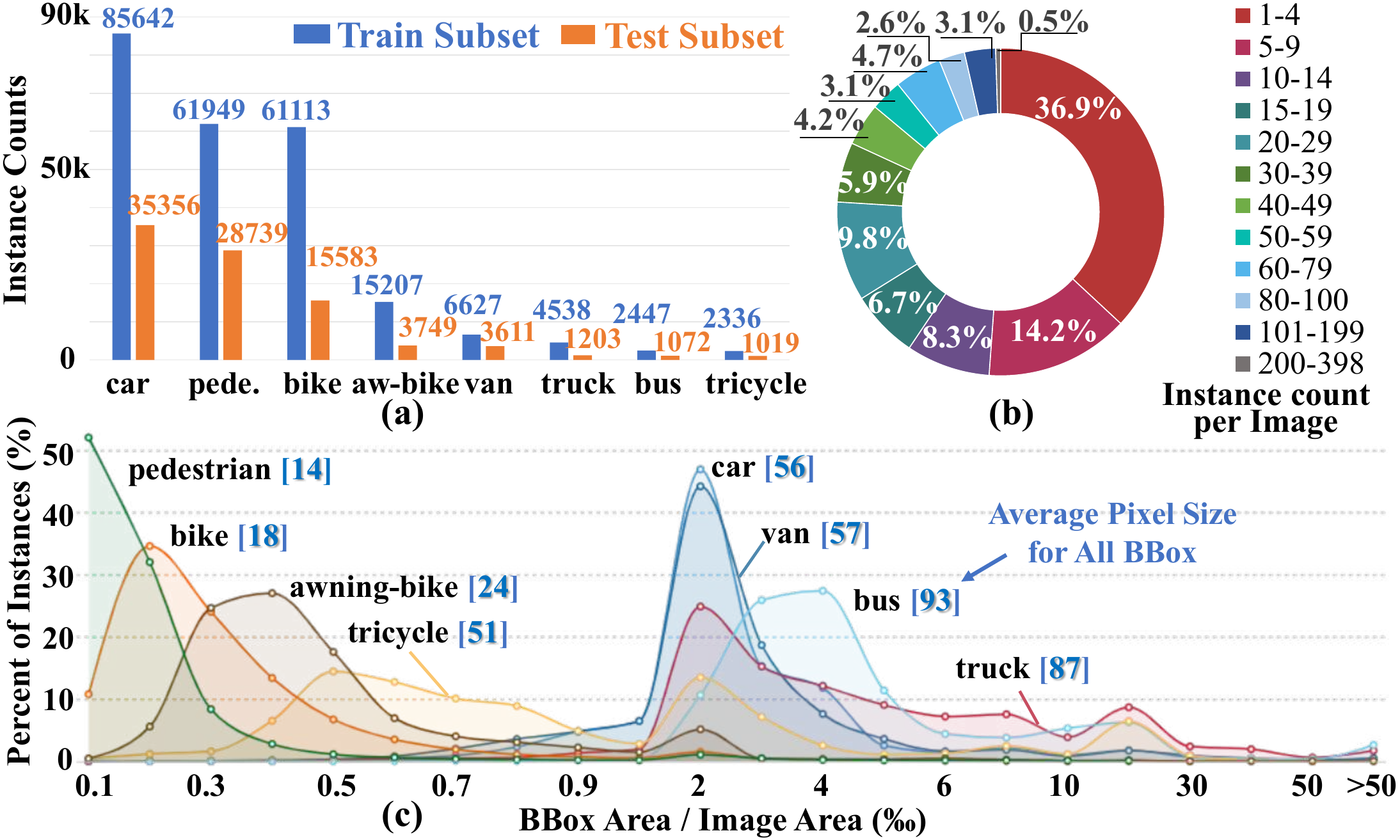}
    \caption{Statistical analysis of MODA. Distribution of instance counts across 8 categories (a) and per MSI (b). (c) Relative and absolute distributions of instance sizes.}
    \label{fig:data_analyse_2}
\end{figure}

\noindent\textbf{Statistical Analysis.} 
MODA comprises 50 scenes, divided into 103 non-overlapping sub-scenes for training (70\%) and testing (30\%), while preserving an approximate category-wise annotation ratio of 7:3, which ensures balanced and distinct challenge attributes and annotations across both subsets (Fig.~\ref{fig:data_analyse_2}~(a)). Fig.~\ref{fig:data_analyse_2}~(b) depicts the instance distribution per MSI, revealing that over 3.5\% of MSIs contain more than 100 instances, contributing to high-density challenges. Furthermore, Fig.~\ref{fig:data_analyse_2}~(c) presents the relative and absolute instance size distributions, indicating that 95\% of instances occupy less than 1\% of the image area. Notably, ``Pedestrian" and ``bike" are particularly small, with an average pixel size below 20, and exhibit significant scale variations.

\begin{figure*}[t]
    \centering
    \includegraphics[width=\linewidth]{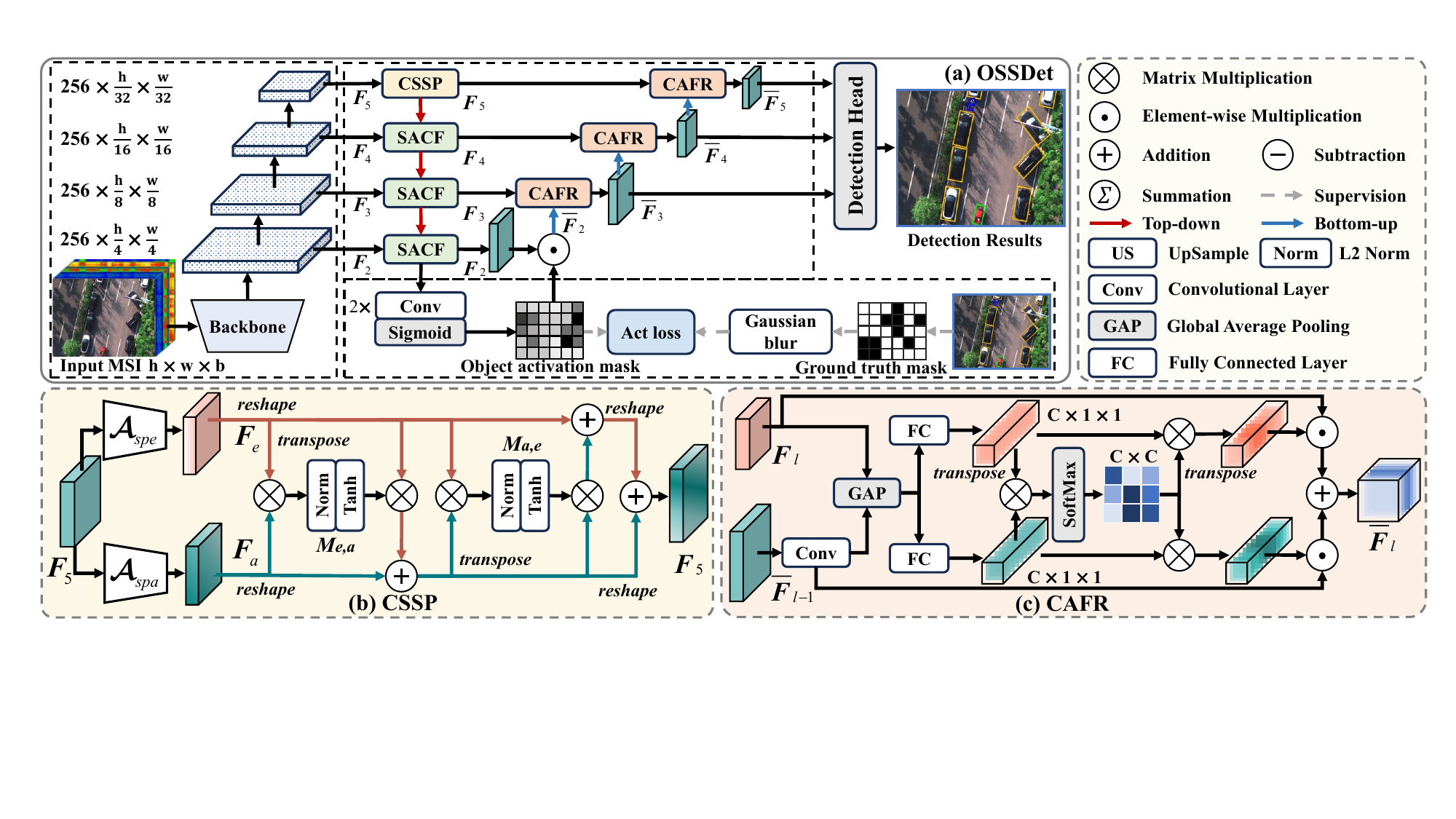}
    \caption{(a) Overall OSSDet framework. SACF fuses aggregated spectral features with spatially enhanced details to reinforce intra-object correlations and spatial texture details; (b) CSSP integrates spectral and spatial awareness to improve target perception; (c) CAFR refines object-related representations with explicit object-aware cues and cross-spectral attention.}
    \label{fig: framework}
\end{figure*}

\section{Method}
We present OSSDet, an object-aware spectral-spatial learning framework for multispectral aerial object detection, as shown in Fig.~\ref{fig: framework}. OSSDet employs Cascaded Spectral-Spatial joint Perception (CSSP) to optimize target region perception through cascaded spectral-spatial awareness interactive modulation. Subsequently, Spectral-guided Adaptive Cross-layer Fusion (SACF) modules build a top-down pathway to propagate target-aware cues across layers, reinforcing intra-object correlations and texture details. Object-aware masking branch filters background noise while retaining discriminative object representations. Finally, Cross-spectral Attention Feature Refinement (CAFR) refines object-related features using object-aware cues and cross-spectral attention.

\subsection{Cascaded Spectral-Spatial Joint Perception}
Successive downsampling during top-level feature extraction introduces spatial aliasing, allowing background clutter to interfere with target regions and degrade model attention. To address this and leverage spectral information, we propose CSSP, which integrates spectral and spatial awareness.

Let $\left\{\textbf{\emph{F}}_{i}\mid i\in{1, ..., 5}\right\}$ denote multiscale features extracted from MSI input $\textbf{\emph{I}}\in\mathbb{R}^{\rm{h}\times w\times b}$, $\rm{b}$ is the bands number. As Fig.~\ref{fig: framework} (b) shows, CSSP first applies spectral and spatial attention to top-level feature $\textbf{\emph{F}}_{5}\in\mathbb{R}^{\rm{C}\times H\times W}$ via two dedicated sub-networks, extracting spectral and spatial aware features.
For spectral-wise attention, $\textbf{\emph{F}}_{5}$ is compacted using global average pooling (GAP), then scaled with learnable weight $\boldsymbol{w}_{e}$ and bias vector $\boldsymbol{b}_{e}$ along the spectral dimension:
\[
 \textbf{\emph{F}}_{e} = \boldsymbol{f}_{s}\left(\boldsymbol{w}_{e} \cdot \rm{GAP}\left(\textbf{\emph{F}}_{5}\right) + \boldsymbol{b}_{e}\right) \odot \textbf{\emph{F}}_{5}, \tag{1}
\]
where $\boldsymbol{f}_{s}\left(\cdot\right)$ is sigmoid, $\odot$ is element-wise multiplication.
Similarly, for spatial-wise attention, $\textbf{\emph{F}}_{5}$ is averaged along the spectral dimension, and a learnable weight map $\boldsymbol{w}_{a}$ and a bias map $\boldsymbol{b}_{a}$ is applied to amplify spatially salient regions:
\[
 \textbf{\emph{F}}_{a} = \boldsymbol{f}_{s}\left(\boldsymbol{w}_{a} \cdot \rm{AVG}\left(\textbf{\emph{F}}_{5}\right) + \boldsymbol{b}_{a}\right) \odot \textbf{\emph{F}}_{5}, \tag{2}
\]

We employ a cascaded interaction modulation network to achieve joint spectral–spatial perception. Specifically, spectral and spatial aware features are reshaped to $\textbf{\emph{F}}_{e}, \textbf{\emph{F}}_{a}\in \mathbb{R}^{\rm{C} \times HW}$, from which the spectral-spatial cross-correlation matrix $\boldsymbol{M}_{e, a}$ is computed to project spectral-aware feature into spatial space, modulating the spatially aware feature:
\[
 \boldsymbol{M}_{e,a} = \mathrm{Tanh}\Big(\frac{\textbf{\emph{F}}_{e}^{T}\textbf{\emph{F}}_{a}}{\big\|\textbf{\emph{F}}_{e}^{T}\textbf{\emph{F}}_{a}\big\|_{2}}\Big), \hat{\textbf{\emph{F}}}_{a} = \textbf{\emph{F}}_{e}\boldsymbol{M}_{e, a} + \textbf{\emph{F}}_{a}, \tag{3}
\]
where $\|\cdot\|_{2}$ denotes the L2 norm. Then, a similar operation is applied to reconstruct the spectrally aware feature $\textbf{\emph{F}}_{e}$ using the modulated spatially aware feature $\hat{\textbf{\emph{F}}}_{a}$:
\[
 \boldsymbol{M}_{a,e} = \boldsymbol{f}_{t}\Big(\frac{\hat{\textbf{\emph{F}}}^{T}_{a}\textbf{\emph{F}}_{e}}{\big\|\hat{\textbf{\emph{F}}}^{T}_{a}\textbf{\emph{F}}_{e}\big\|_{2}}\Big), \hat{\textbf{\emph{F}}}_{e} = \hat{\textbf{\emph{F}}}_{a}\boldsymbol{M}_{a,e} + \textbf{\emph{F}}_{e}, \tag{4}
\]

Finally, the modulated spectral and spatial features are reshaped back and fused to form the reconstructed feature $\hat{\textbf{\emph{F}}}_{5}$.

\subsection{Spectral-guided Adaptive Cross-layer Fusion}
\begin{figure}[t]
    \centering
    \includegraphics[width=\linewidth]{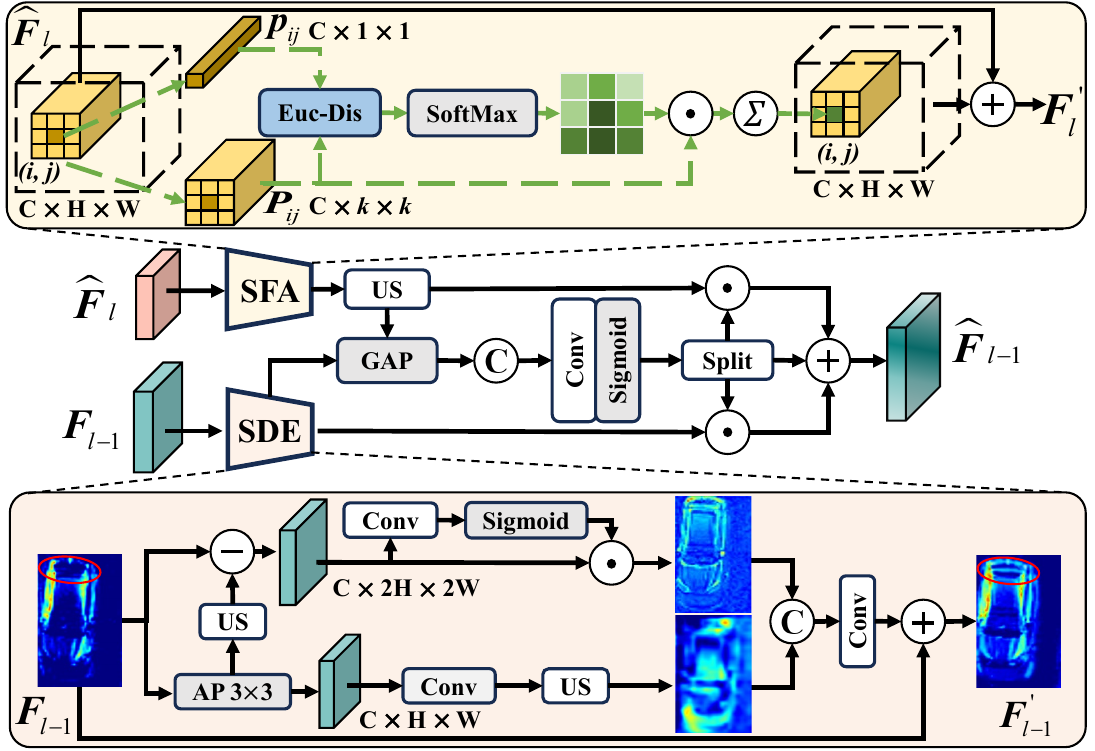}
    \caption{Illustration of SACF. The Euc-Dis denotes Euclidean Distance for spectral vector similarity measurement.}
    \label{fig:SACF}
\end{figure}

As shown in Fig.~\ref{fig:SACF}, SACF incorporates Spectral Feature Aggregator (SFA) to adaptively aggregate spectral information and Spatial Detail Enhancer (SDE) to improve spatial details in high-resolution features. A cross-layer fusion unit subsequently embeds the aggregated spectral features and enhanced spatial textures into the low-level feature.

\noindent\textbf{Spectral Feature Aggregator.} 
SFA adaptively aggregates spectral features by examining the relationship between a central spectral feature vector $\boldsymbol{p}_{i,j} \in \mathbb{R}^{\rm C \times 1 \times 1}$ and its neighbors within a $k \times k$ spatial patch $\boldsymbol{P}_{i, j}$ centered at $(i, j)$ in the high-level feature $\hat{\boldsymbol{F}}_{l}\in\mathbb{R}^{\rm{C}\times H\times W}$, where $\boldsymbol{P}_{i,j} = \left[ \boldsymbol{p}_{i+\delta_m, j+\delta_n} \right]_{\delta_m, \delta_n}$, 
and $-\frac{k-1}{2} \leq \delta_m, \delta_n \leq \frac{k-1}{2}$ denotes the spatial offsets. To quantify the similarity between the central spectral feature vector \( \boldsymbol{p}_{i,j} \) and its neighbors, the similarity weight \( e_{i+\delta_m, j+\delta_n} \) is computed for each neighboring vector:
\[
e_{i+\delta_m, j+\delta_n} = \frac{\exp\left(-\| \boldsymbol{p}_{i,j} - \boldsymbol{p}_{i+\delta_m, j+\delta_n} \|_2\right)}{\sum_{\delta_u, \delta_v} \exp\left(-\| \boldsymbol{p}_{i,j} - \boldsymbol{p}_{i+\delta_u, j+\delta_v} \|_2\right)}, \tag{5}
\]
where \( \delta_u, \delta_v \) share same spatial offsets with \( \delta_m, \delta_n \). Then the central spectral feature vector \( \boldsymbol{p}_{i,j} \) is updated by weighted aggregation of the spectral feature vectors within \( \boldsymbol{P}_{i,j} \):
{
\begin{equation}
\begin{array}{c}
\boldsymbol{p}'_{i,j} = \sum_{\delta_m, \delta_n} e_{i+\delta_m, j+\delta_n} \cdot \boldsymbol{p}_{i+\delta_m, j+\delta_n} + \boldsymbol{p}_{i,j}, 
\end{array}
\tag{6}
\end{equation}
}

Finally, \( \hat{\boldsymbol{F}}_{l} \) is updated to \( \boldsymbol{F}'_l \) by adaptively aggregating neighboring spectral information for each spectral vector, which enhances intra-object feature correlations, thereby improving robustness against noise interference.

\noindent\textbf{Spatial Detail Enhancer.} 
Downsampling during feature extraction can induce spectral aliasing, which destabilizes feature aggregation—particularly at edges with prevalent mixed pixels—and leads to a loss of spatial texture details. To mitigate this, the SDE is introduced to enhance spatial details, minimize information loss during cross-layer transmission.

For a low-level feature $\boldsymbol{F}_{l-1}\in\mathbb{R}^{\rm{C} \times 2H\times 2W}$, SDE first decomposes it into low-frequency $\boldsymbol{F}_{l-1}^{lf}\in\mathbb{R}^{\rm C\times H\times W}$ and high-frequency $\boldsymbol{F}_{l-1}^{hf}\in\mathbb{R}^{\rm C\times 2H\times 2W}$ components: 
{
\begin{equation}
\begin{array}{c}
\boldsymbol{F}_{l-1}^{lf} = \mathrm{AP}(\boldsymbol{F}_{l-1}), \boldsymbol{F}_{l-1}^{hf} = \boldsymbol{F}_{l-1} - \mathrm{US}(\boldsymbol{F}_{l-1}^{lf}), 
\end{array}
\tag{7}
\end{equation}
}

\noindent where \(\mathrm{AP}\) is $3\times 3$ average pooling. Next, the high-frequency component, which stores spatial texture details (Fig.~\ref{fig:SACF}), is enhanced via a lightweight detail enhancement block: 
\[
\hat{\boldsymbol{F}}_{l-1}^{hf} = (1 + \boldsymbol{f}_{s}(\boldsymbol{\phi}_{1\times 1}(\boldsymbol{F}_{l-1}^{hf}))) \odot \boldsymbol{F}_{l-1}^{hf}, \tag{8}
\]
where $\boldsymbol{\phi}_{k\times k}$ defines a $k \times k$ convolution. Finally, the detail-enhanced feature is obtained by fusing the enhanced high-frequency textures with low-frequency structural context:
\[
\boldsymbol{F}'_{l-1} = \boldsymbol{\phi}_{1\times 1}([\hat{\boldsymbol{F}}_{l-1}^{hf}, \mathrm{US}(\boldsymbol{\phi}_{3\times 3}(\boldsymbol{F}_{l-1}^{lf}))]) + \boldsymbol{F}_{l-1}, \tag{9}
\]
where $[\cdot]$ denotes channel-wise concatenation.

\noindent\textbf{Adaptive Cross-layer Fusion.} 
To adaptively fuse aggregated spectral features with enhanced spatial details, we first align the spatial dimensions of $\boldsymbol{F}'_{l}$ and $\boldsymbol{F}'_{l-1}$. Global confidence are then computed by applying GAP to them separately, which are concatenated and passed a $1\times1$ convolution followed by activation to generate adaptive fusion weights:
\begin{equation}
    \begin{array}{c}
        \boldsymbol{W} = \boldsymbol{f}_{s}(\boldsymbol{\phi}_{1\times 1}([\mathrm{AVG}(\boldsymbol{F}'_{l}), \mathrm{AVG}(\boldsymbol{F}'_{l-1})])), \\[1.5pt]
        \boldsymbol{W}_{l}, \boldsymbol{W}_{l-1} = \boldsymbol{f}_{split}(\boldsymbol{W})\in\mathbb{R}^{\rm 2H\times2W},
    \end{array}
    \tag{10}
\end{equation}
where $\boldsymbol{f}_{split}(\cdot)$ denotes channel-wise split. Finally, we embed $\boldsymbol{F}'_{l}$ and $\boldsymbol{F}'_{l-1}$ to low-level features via adaptive weighted summation: $\hat{\boldsymbol{F}}_{l-1} = \boldsymbol{F}'_{l} \odot \boldsymbol{W}_{l} + \boldsymbol{F}'_{l-1} \odot \boldsymbol{W}_{l-1}$, thereby increasing foreground-background contrast.

\subsection{Object-aware Masking} 
\label{Object-aware Masking}
SACFs propagate target-aware information across network stages. However, preserving distinctive features for instances prone to background interference, such as small objects, remains challenging. To address this, we introduce the Object Activation Loss $\mathcal{L}_{act}$, which explicitly enforces object-specific activations.
As shown in Fig.~\ref{fig: framework} (a), the spectral-spatial features of objects in $\hat{\boldsymbol{F}}_{2}$ is projected to an activation representations: $\boldsymbol{M}_{p} = \boldsymbol{f}_{s}(\boldsymbol{\phi}_{3\times 3}(\boldsymbol{\phi}_{3\times 3}(\hat{\boldsymbol{F}}_{2})))$, 
where $\boldsymbol{M}_{p}\in\mathbb{R}^{\rm H \times W}$ is the predicted object activation mask. 

Simultaneously, the ground truth instance mask $\boldsymbol{M}_{g} \in \mathbb{R}^{\rm H \times W}$ is generated from bounding boxes using a Gaussian blur to model the instance as 2D Gaussian distribution.

$\mathcal{L}_{act}$ comprises an intersection loss $\mathcal{L}_{I}$ promotes the activation of target-related features, and a difference loss $\mathcal{L}_{D}$ penalizes the activation of irrelevant background features.
{
\small
\[
\mathcal{L}_{I}=1-\frac{\textstyle \sum \boldsymbol{M}_p \boldsymbol{M}_g}{\textstyle \sum \boldsymbol{M}_g}, \quad
\mathcal{L}_{D}=\frac{\textstyle \sum \boldsymbol{M}_p (1 - \mathcal{H}(\boldsymbol{M}_g))}{\textstyle \sum \boldsymbol{M}_p},
\tag{11}
\]
}

\noindent where $\mathcal{H}(\cdot)$ maps values greater than 0 to 1. Finally, the object activation loss is formulated as: $\mathcal{L}_{act} = \mathcal{L}_{I} + \gamma \mathcal{L}_{D}$, 
with $\gamma$ controlling foreground-background balance.

Supervision by $\mathcal{L}_{act}$, $\boldsymbol{M}_{p}$ highlights target regions, enabling  $\hat{\boldsymbol{F}}_{2}$ to be masked as $\overline{\boldsymbol{F}}_{2} = \hat{\boldsymbol{F}}_{2} \odot \boldsymbol{M}_{p}$ to retain object-specific features while suppressing irrelevant response.

\subsection{Cross-spectral Attention Feature Refinement}
To further refine object-related representations in multiscale features using explicit object-aware cues from $\overline{\boldsymbol{F}}_{2}$, we construct a bottom-up flow with CAFRs. As shown in Fig.~\ref{fig: framework} (c), given $\hat{\boldsymbol{F}}_{l} \in \mathbb{R}^{\rm C \times H \times W}$ and its lower-level feature $\overline{\boldsymbol{F}}_{l-1} \in \mathbb{R}^{\rm C \times 2H \times 2W}$, CAFR first aligns them via a $3\times3$ convolution.
Global average pooling followed by a fully connected layer yields global spectral vectors $\boldsymbol{V}_{l}, \boldsymbol{V}_{l-1} \in \mathbb{R}^{\rm C \times 1 \times 1}$, which are used to calculate fusion weights $\boldsymbol{W}_{l}, \boldsymbol{W}_{l-1}$ through cross attention, mitigating spectral bias across layers:

\begin{equation}
    \begin{array}{c}
        \boldsymbol{A} = softmax(\boldsymbol{V}_{l}^{T}\boldsymbol{V}_{l-1}), \\[1.5pt]
        \boldsymbol{W}_{l}, \boldsymbol{W}_{l-1} = \boldsymbol{A}^{T}\boldsymbol{V}_{l}, \boldsymbol{A}\boldsymbol{V}_{l-1},
    \end{array}
    \tag{12}
\end{equation}

Finally, $\hat{\boldsymbol{F}}_{l}$ is refined by embedding spatial object-aware and aligned spectral cues: $\overline{\boldsymbol{F}}_{l} = \boldsymbol{W}_{l} \odot \hat{\boldsymbol{F}}_{l} + \boldsymbol{W}_{l-1} \odot \overline{\boldsymbol{F}}_{l-1}$.

\subsection{Learning Objective}
OSSDet incorporates detection loss $\mathcal{L}_{det}$ from \cite{oriented-reppoints} and $\mathcal{L}_{act}$ introduced above, the total model optimize function is:
\[
\mathcal{L} = \mathcal{L}_{det} + \alpha \mathcal{L}_{act}.  \tag{13}
\]
where $\alpha$ is a weighting factor.

\begin{table*}[!ht]   
\centering
\small
{
\setlength{\tabcolsep}{2.0pt}
\begin{tabular}{l|l|cccccccc|ccc|cc} 
\toprule
\multicolumn{2}{c|}{\textbf{Method}}                                                      & \textbf{car}  & \textbf{bus}  & \textbf{van}  & \textbf{aw-bike} & \textbf{truck} & \textbf{tricycle} & \textbf{bike} & \textbf{pedestrian} & \textbf{mAP$_{50}$} & \textbf{mAP$_{75}$} & \textbf{mAP}  & \textbf{FLOPs} & \textbf{Params}  \\ 
\midrule
\multirow{3}{*}{\scriptsize \rotatebox{90}{\textbf{Two-stage}}}  & Gliding Vertex \cite{Gliding-vertex}     & 90.3          & 89.1          & 73.8          & 69.5             & 66.0           & 46.7              & 41.4          & 22.6                & 62.4                & 35.7                & 34.7          & 230.8G          & 41.4M                \\
                                               & Roi Transformer \cite{roi_trans}         & \textbf{90.5} & 89.3          & 75.2          & 73.3    & 68.7           & 51.8              & 44.5          & 30.0                & 65.4                & 43.4                & 40.7          & 244.7G          & 55.3M                \\
                                               & StripRCNN \cite{yuan2025strip}         & \textbf{90.5} & 89.0          & 76.1          & \textbf{73.6}             & 67.1           & 56.5              & 45.2          & 30.7                & 66.1                & 44.0                & 41.0          & 231.8G          & 45.2M                \\ 
\midrule
\multirow{10}{*}{\scriptsize \rotatebox{90}{\textbf{One-stage}}} 
                                               & GWD \cite{yang2021rethinking}            & 90.4          & 79.9          & 70.7          & 59.7             & 49.3           & 30.8              & 26.7          & 29.7                & 54.7                & 37.8                & 34.6          & 233.5G          & 36.5M                \\
                                               & R3Det \cite{yang2021r3det}               & 90.4          & 88.7          & 74.0          & 63.8             & 57.8           & 45.1              & 29.4          & 32.2                & 60.2                & 36.7                & 34.8          & 362.2G          & 42.0M                \\
                                               & S$^2$ANet \cite{han2021align}            & 90.4          & 88.7          & 74.4          & 69.1             & 62.7           & 48.9              & 30.1          & 40.7                & 63.1                & 38.7                & 36.7          & 216.5G          & 38.8M                \\
                                               & R3Det-KLD \cite{yang2021learning}        & 90.4          & 88.7          & 74.8          & 66.6             & 60.9           & 42.6              & 29.8          & 35.5                & 61.1                & 39.4                & 36.8          & 306.7G          & 39.6M                \\
                                               & LSKNet-S$^2$ANet \cite{Li_2024_IJCV}       & \textbf{90.5}          & 89.5          & 76.1          & 72.5             & 68.6           & 51.8              & 38.9          & 43.1                & 66.4                & 41.3                & 38.6          & \textbf{194.9G}          & \textbf{29.9M}               \\
                                               & S2ADet$^\dagger$  \cite{he2023object}               & 90.3          & 86.6          & 72.1          & 71.3             & 57.2           & 54.1              & 35.0          & 40.8                & 63.5                & 41.1                & 38.9          & 406.0G          & 65.2M               \\
                                               & Rot. FCOS \cite{tian2019fcos}            & 90.3          & 86.3          & 75.2          & 71.2             & 59.0           & 53.8              & 34.7          & 37.4                & 63.5                & 41.8                & 39.1          & 226.9G          & 32.2M       \\
                                               & CFA \cite{cfa}                           & 90.4          & 89.0          & 76.7          & 69.7             & 64.4           & 55.1              & 43.1          & 41.5                & 66.2                & 43.2                & 40.6          & 213.6G & 36.9M                \\
                                               & Ori. RepPoints \cite{oriented-reppoints} & \textbf{90.5} & 89.2          & 77.7          & 71.2             & 66.2           & 53.1              & 43.0          & 41.1                & 66.5                & 44.1                & 40.9          & 213.6G & 36.9M                \\ 
\midrule
\multicolumn{1}{l}{}                           & \textbf{OSSDet (Ours)}                   & \textbf{90.5} & \textbf{89.9} & \textbf{79.2} & 72.7             & \textbf{69.7}  & \textbf{58.8}     & \textbf{45.3} & \textbf{45.7}       & \textbf{69.0}       & \textbf{45.9}       & \textbf{42.7} & 263.1G          & 36.5M                \\
\bottomrule
\end{tabular}
}
\caption{Comparison with other methods on MODA. $^\dagger$ indicates methods originally designed for multispectral object detection.}
\label{tab2}
\end{table*}

\section{Experiments}
We evaluate OSSDet on MODA and HOD3K. For fairness, all methods are evaluated in identical settings. Further results and details are available in the supplementary material.

\subsection{Results on the MODA Dataset}
\noindent \textbf{Quantitative Results.} Table~\ref{tab2} reports results on MODA. Overall, OSSDet achieves superior performance, outperforming the suboptimal method by 2.5\%, 1.8\%, and 1.7\% in mAP${50}$, mAP${75}$, and mAP, respectively. The latest MSI-oriented detector S2ADet struggles with small objects like ``bike” and ``pedestrian”. OSSDet achieves top results in all categories except ``awning-bike", with notable gains of 2.3\% and 2.6\% in ``tricycle" and ``pedestrian", underscoring its effectiveness in preserving small objects' distinctive features.

\begin{table}[t]
    \centering 
    \small
    \setlength{\tabcolsep}{0.8pt}
        \begin{tabular}{l|c c c|c c}
            \toprule
            Method & mAP$_{50}$ & mAP$_{75}$ & mAP & FLOPs & Params \\
            \midrule
            CO-DETR (Zong et al. 2023) & 77.9 & 57.3 & 51.7 & 200.7G & 64.5M \\ 
            DINO \cite{zhang2022dino} & 89.1 & 61.6 & 56.3 & 114.7G & 47.6M \\ 
            Fovea \cite{foveabox} & 92.2 & 62.2 & 56.8 & 123.9G & 38.0M \\
            TOOD \cite{tood} & 90.3 & 65.6 & 57.8 & 108.5G & \textbf{32.1M}\\
            VFNet \cite{zhang2021varifocalnet} & 92.3 & 66.5 & 59.0 & \textbf{104.5G} & 32.8M \\
            C-RCNN (Cai et al. 2019) & 90.9 & 66.9 & 59.6 & 149.5G & 69.2M \\
            S2ADet$^\dagger$ \cite{he2023object}& \textbf{93.4} & - & 59.8 & 169.2G & 48.6M \\
            \midrule
            \textbf{OSSDet (Ours)} & \textbf{93.4} & \textbf{68.8} & \textbf{60.9} & 131.2G & 36.6M \\
            \bottomrule
        \end{tabular}
        \caption{Comparison with other detectors on HOD3K. 
    }
    \label{tab3}
\end{table}

\begin{table}[t]
\centering
\small
    \setlength{\tabcolsep}{0.5pt}
    \begin{tabular}{c|c|ccc|cc} 
\toprule
Method                                                                       & Input & mAP$_{50}$            & mAP$_{75}$            & mAP                 & FLOPs & Params \\ 
\midrule
\multirow{2}{*}{\begin{tabular}[c]{@{}c@{}}Strip\\RCNN\end{tabular}}   & RGB   & 65.2                & 41.9                & 39.6                & 227.4G& 45.13M       \\ 
                                                                             & MSI   & 66.1 (+0.9)          & 44.0 (+2.1)          & 41.0 (+1.4)          & 231.8G & 45.15M     \\ 
\midrule
\multirow{2}{*}{\begin{tabular}[c]{@{}c@{}}Oriented\\RepPoints\end{tabular}} & RGB   & 66.2                & 41.6                & 39.4                & 209.2G  & 36.83M       \\ 
                                                                             & MSI   & 66.5 (+0.3)          & 44.1 (+2.5)          & 40.9 (+1.5)          & 213.6G & 36.85M    \\ 
\midrule
\multirow{2}{*}{\begin{tabular}[c]{@{}c@{}}\textbf{OSSDet}\\\textbf{(Ours)}\end{tabular}}                                             & RGB   & 67.1                & 41.3                & 39.7                & 258.7G & 36.46M      \\ 
                                                                             & MSI   & \textbf{69.0 (+1.9)} & \textbf{45.9 (+4.6)} & \textbf{42.7 (+3.0)} & 263.1G & 36.48M       \\
\bottomrule
    \end{tabular}
\caption{Comparison with different input.}
\label{tab4}
\end{table}
\begin{table}[t]
    \centering
    \small
    \setlength{\tabcolsep}{4.5pt}
    \begin{tabular}{cccc|ccc}
        \toprule
        CSSP & SACF & Obj-aware & CAFR & mAP$_{50}$ & mAP$_{75}$ & mAP \\
        \midrule
        \ding{55} & \ding{55} & \ding{55} & \ding{55} & 66.5 & 44.1 & 40.9 \\
        \checkmark & \ding{55} & \ding{55} & \ding{55} & 67.6 & 44.1 & 41.4 \\
        \ding{55} & \checkmark & \ding{55} & \ding{55} & 67.4 & 44.7 & 41.6 \\
        \checkmark & \checkmark & \ding{55} & \ding{55} & 68.1 & 44.9 & 41.9 \\
        \checkmark & \checkmark & \checkmark & \ding{55} & 68.7 & 45.6 & 42.5 \\
        \checkmark & \checkmark & \checkmark & \checkmark  & \textbf{69.0} & \textbf{45.9} & \textbf{42.7} \\
        \bottomrule
    \end{tabular}
        \caption{Ablation studies on the key components of OSSDet.}
    \label{tab5}
\end{table}
\begin{table}[t]
\centering
\small
\setlength{\tabcolsep}{12pt}
    \begin{tabular}{cc|ccc}
    \toprule
    SDE & SFA & mAP$_{50}$ & mAP$_{75}$ & mAP\\ 
    \midrule
    \checkmark & \ding{55} & 67.8 & 44.5 & 41.5 \\ 
    \ding{55} & \checkmark & 68.2 & 44.7 & 41.7 \\ 
    \checkmark & \checkmark & \textbf{69.0} & \textbf{45.9} & \textbf{42.7} \\ 
    \bottomrule
    \end{tabular}
    \caption{Ablation study of the key components of SACF.}
\label{tab:sacf_ablation}
\end{table}

\noindent \textbf{Efficiency Analysis.} As Table~\ref{tab2} shows, two-stage methods offer high accuracy but complex architectures, while single-stage ones like LSKNet are lightweight yet less accurate. The MSI-oriented S2ADet adopts a two-stream design to decouple spatial and spectral cues, incurs substantial preprocessing and computational cost. In contrast, OSSDet achieves best accuracy with 36.5M parameters and 263.1G FLOPs, offering a favorable accuracy-efficiency balance.

\begin{figure*}
    \centering
    \includegraphics[width=\linewidth]{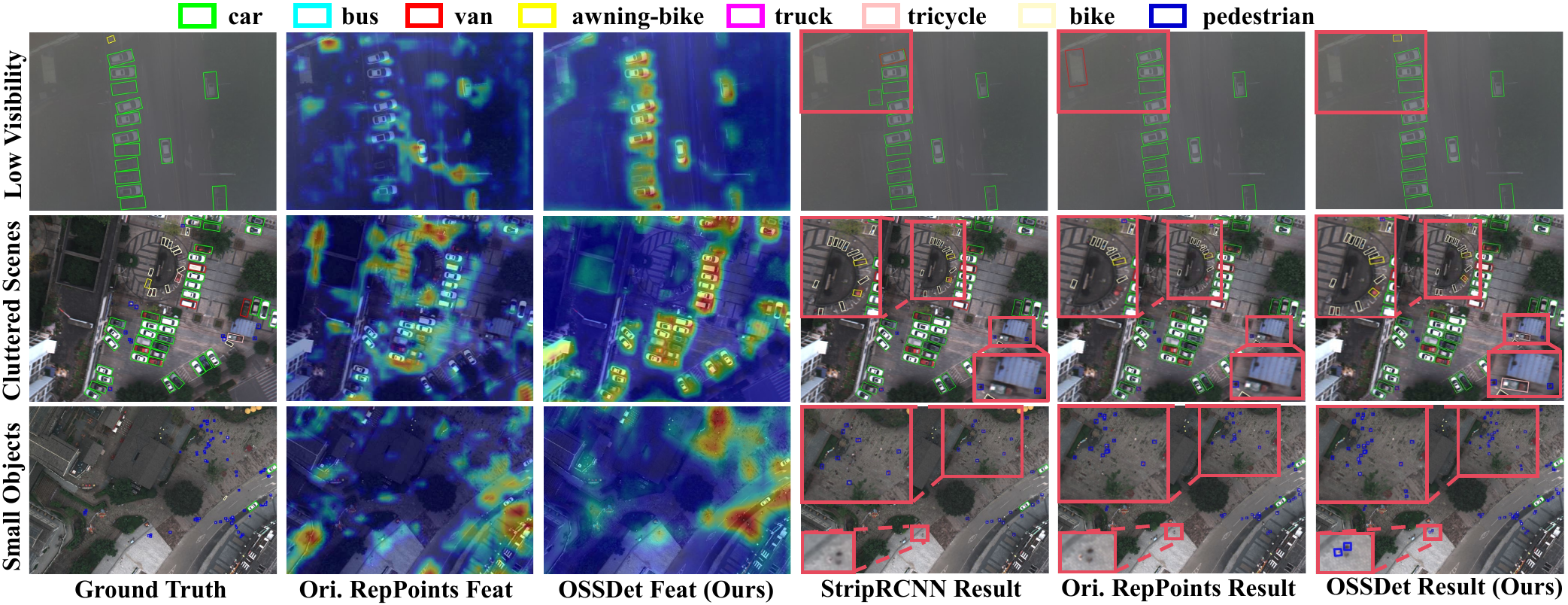}
    \caption{Visualization comparison of detection results and feature maps obtained by different methods on MODA.}
    \label{fig:MODA_results_vis}
\end{figure*}

\noindent\textbf{Qualitative Results.} 
As shown in Fig.~\ref{fig:MODA_results_vis}, OSSDet enhances the network's focus on target regions while suppressing irrelevant noise, even for challenging scenarios, such as small objects, low visibility, or cluttered backgrounds, resulting in a substantial reduction of false and missed detections.

\begin{figure}[t]
    \centering
    \includegraphics[width=\linewidth]{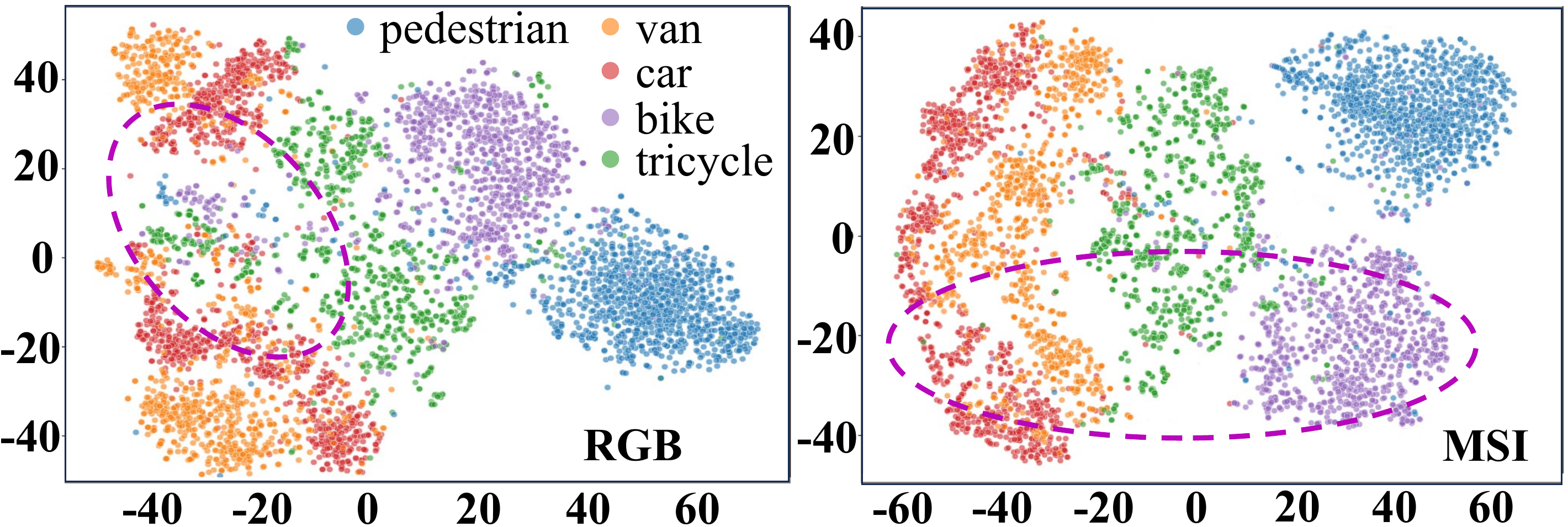}
    \caption{2D t-SNE visualization of feature distributions learned by OSSDet on MODA with RGB and MSI inputs.}
    \label{fig:tsne}
\end{figure}

\noindent\textbf{Visualization of Feature Distribution.}
Fig.~\ref{fig:tsne} shows T-SNE~\cite{van2008visualizing} visualization, where MSI inputs enlarge inter-class distances for easily confused (car \& van), small (pedestrian \& bike), and few-sample (tricycle) categories, yielding clearer feature separation. This confirms that spectral cues improve object discriminability and validate OSSDet’s effectiveness in feature learning.

\begin{figure}[t]
    \centering
    \includegraphics[width=\linewidth]{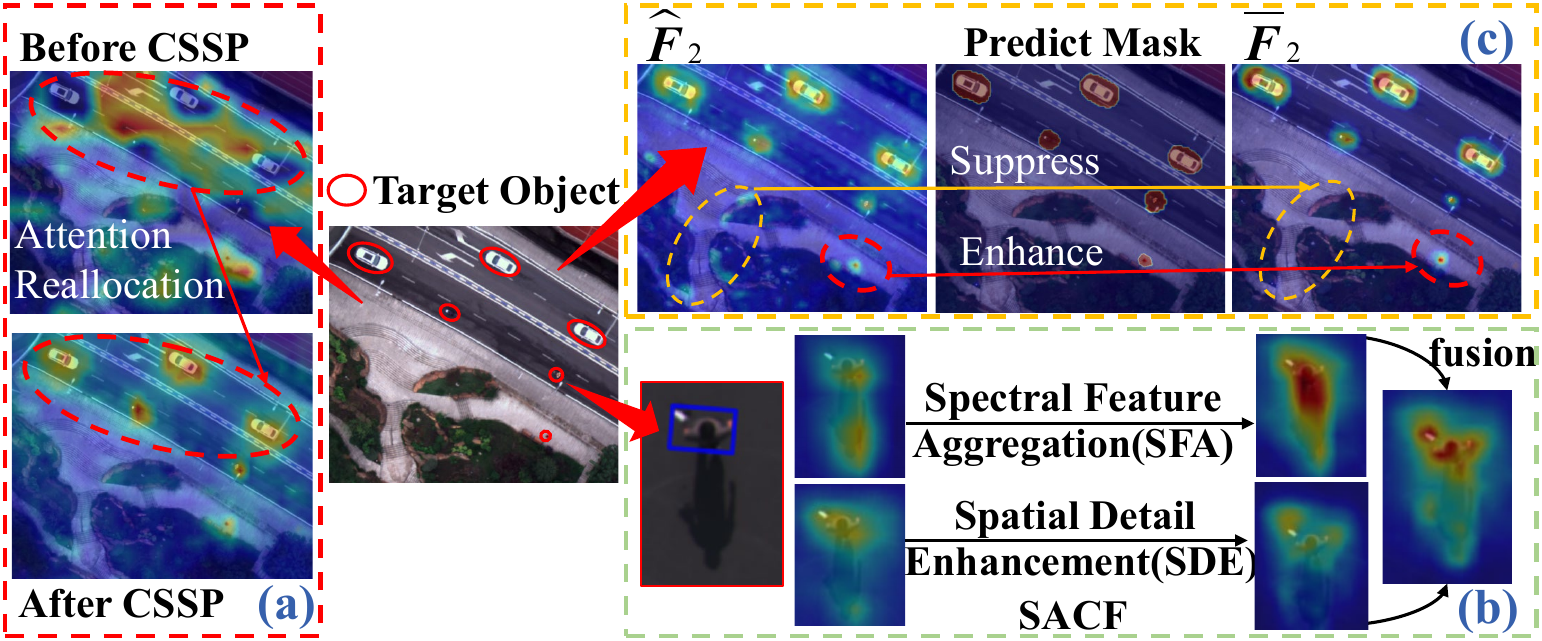}
    \caption{Effectiveness visualization of key components.}
    \label{fig:features}
\end{figure}

\noindent\textbf{Effect of Components.} Fig.~\ref{fig:features} visualizes features involved in key components. (a) CSSP optimizes target perception via cascaded spectral–spatial modulation; (b) SACF integrates spectral aggregation with spatial detail enhancement to strengthen intra-object correlations and clear boundaries; (c) object-aware masking backpropagates $\mathcal{L}_{act}$ to preserve object features, even for small objects-reinforces object-specific features while suppressing noise. 

\subsection{Results on the HOD3K Dataset}
\noindent\textbf{Quantitative Results.}  
OSSDet achieves the best overall performance, surpassing the suboptimal method by 1.9\% in mAP${75}$ and 1.1\% in mAP, as shown in Table~\ref{tab3}. While S2ADet matches OSSDet in mAP${50}$, its two-stream design causes high preprocessing and computation overhead. In contrast, OSSDet directly processes MSI inputs, delivering better accuracy with fewer parameters and FLOPs.

\begin{figure}[!ht]
    \centering
    \includegraphics[width=\linewidth]{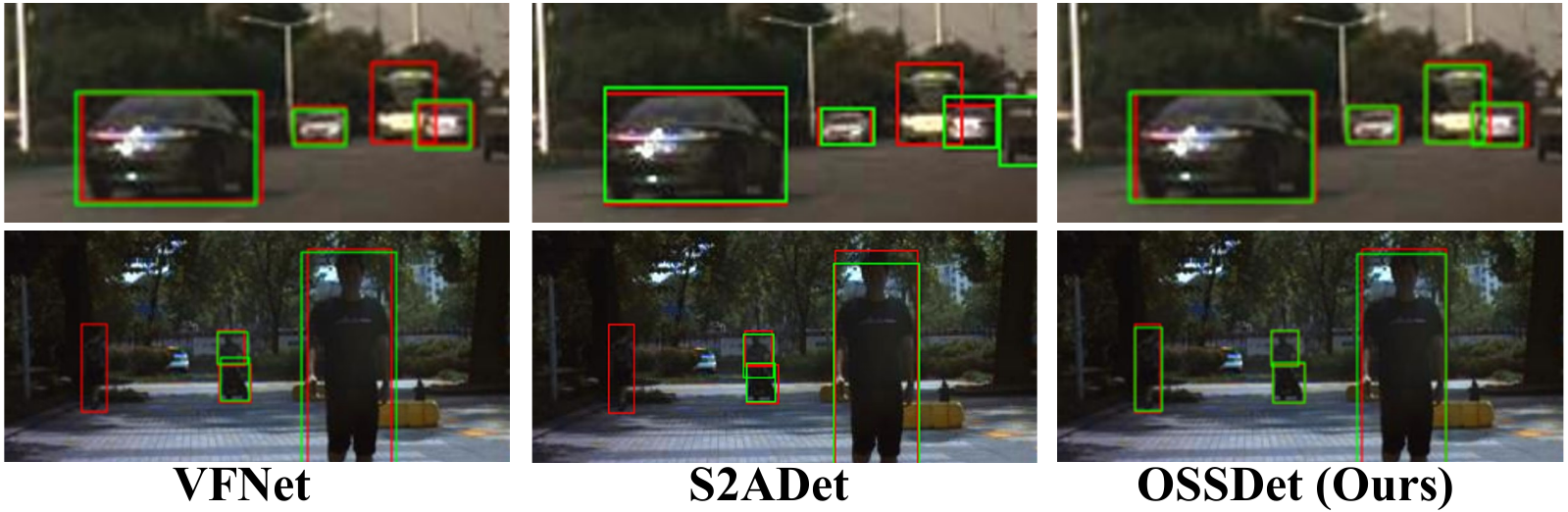}
    \caption{Detection results on HOD3K. \textbf{Red} boxes indicate the ground truth, while \textbf{green} boxes denote the predictions.}
    \label{fig:hod3k_result_vis}
\end{figure}

\noindent\textbf{Qualitative Results.}  
Other methods often miss occluded or background-blended objects (Fig.~\ref{fig:hod3k_result_vis}). In contrast, OSSDet exploits spectral cues to enhance target–background separation, yielding predictions closer to the ground truth.


\subsection{Ablation Studies}
\textbf{Input Ablation Study.} 
As shown in Table~\ref{tab4}, MSI inputs significantly improve performance over RGB due to richer spectral information. OSSDet achieves the largest gains by effectively integrating spectral–spatial features with object-aware cues, while adding only marginal cost (4.4 GFLOPs, 0.02 M params), making this enhancement nearly cost-free.

\noindent\textbf{Component Ablation Study.} 
Each component improves detection, with full integration performing best (Table~\ref{tab5}). Notably, adding SACF and object-aware branch offers the most significant gains across metrics, underscoring the value of integrating spectral aggregation, spatial detail enhancement, and object-aware cues for spectral–spatial learning.

\noindent\textbf{Ablation Study on SACF.} 
As shown in Table~\ref{tab:sacf_ablation}, removing either SACF component causes clear performance drops, confirming their complementarity. SACF adaptively aggregates spectral features within a spatial patch of size $k$, overly large patches introduce irrelevant features that weaken target specificity and degrade performance (Table~\ref{tab:cssp_k}).

\begin{table}[t]
\centering
\small
\setlength{\tabcolsep}{3.5pt}
    \begin{tabular}{c|ccc||c|ccc}
        \toprule
        Method & mAP$_{50}$ & mAP$_{75}$ & mAP & $k$ & mAP$_{50}$ & mAP$_{75}$ & mAP\\
        \midrule
        SoftMax & 66.7 & 42.8 & 40.5 & 3 & \textbf{69.0} & \textbf{45.9} & \textbf{42.7}\\
        Concat & 67.9 & 44.7 & 41.9 & 5 & 68.7 & 45.8 & 42.4 \\
        Addition & 68.5 & 45.4 & 42.4 & 7 & 61.6 & 40.4 & 37.6 \\
        \textbf{CSSP} & \textbf{69.0} & \textbf{45.9} & \textbf{42.7} & 9 & 61.4 & 38.5 & 36.8\\
        \bottomrule
    \end{tabular}
    \caption{Ablation study of CSSP and patch size $k$ in SACF.}
\label{tab:cssp_k}
\end{table}

\begin{table}[t]
\centering
\small
\setlength{\tabcolsep}{4.2pt}
    \begin{tabular}{c|ccc||c|ccc}
        \toprule
        $\boldsymbol{\mathbf{\gamma}}$ & mAP$_{50}$ & mAP$_{75}$ & mAP & $\boldsymbol{\alpha}$ & mAP$_{50}$ & mAP$_{75}$ & mAP\\
        \midrule
        0.5 & 67.6 & 44.8 & 41.8 & 0.4 & 68.2 & 45.2 & 41.8\\
        0.25 & 68.2 & 45.4 & 42.1 & \textbf{0.6} & \textbf{69.0} & \textbf{45.9} & \textbf{42.7} \\
        \textbf{0.1} & \textbf{69.0} & \textbf{45.9} & \textbf{42.7} & 0.8 & 68.0 & 44.6 & 41.7 \\
        0.05 & 67.4 & 44.6 & 41.5 & 1.0 & 68.3 & 45.1 & 42.0\\
        \bottomrule
    \end{tabular}
    \caption{Ablation study of weighting factors.}
\label{tab:weighting_factors}
\end{table}

\noindent\textbf{Ablation Study on CSSP.} 
CSSP enhances target perception through cascaded spectral–spatial interaction. To assess its effectiveness, we compare it with addition, concatenation, and a Softmax-based variant (Table~\ref{tab:cssp_k}), CSSP outperforms all, while the Softmax version shows unstable convergence.

\noindent\textbf{Ablation Study on Weighting Factors.} 
We evaluated the effects of $\gamma$ and loss weight $\alpha$ (Table~\ref{tab:weighting_factors}). A large $\gamma$ suppresses early activations, whereas a small one fails to filter noise. OSSDet performs best with $\gamma=0.1$ and $\alpha=0.6$.

\section{Conclusion}
We introduce MODA, the first large-scale challenging dataset for multispectral aerial object detection, which addresses training sample scarcity with high-resolution MSIs featuring diverse scenarios, challenging attributes, and high-quality oriented annotations. We also propose OSSDet, a novel baseline integrating spectral-spatial information with object-aware cues. OSSDet enhances target perception via cascaded spectral-spatial interactive modulation, reinforces intra-object correlations via spectral aggregation, and suppresses background noise through explicit object-aware guidance, boosting detection accuracy. We believe MODA and OSSDet will catalyze future research in this domain.


\bibliography{aaai2026}

\end{document}